# Neuro-Bayesian-Symbolic Residual Attention Shallow Network: Explainable Deep Learning for Cybersecurity Risk Assessment

Nicolaie Popescu-Bodorin*, Madeleine Togher*

**Nicolaie Popescu-Bodorin***
Mathematics & Computer Science Department, Ovidius University, Constanta, Romania
Fundamental Sciences & Humanities Department, Maritime University, Constanta, Romania
Engineering & Computer Science Department, Spiru Haret University, Bucharest, Romania
*Corresponding author: bodorin@live.com

**Madeleine Togher***
Computer Information Science Department, Higher Colleges Technologies, Sharjah, U.A.E.
*Corresponding author: mtogher@hct.ac.ae

**Abstract**

We introduce a novel hybrid neural architecture – the Neuro-Bayesian-Symbolic Residual Attention Shallow Network (NBS-RASN) – for explainable cybersecurity risk evaluation in open-source software ecosystems. Unlike conventional deep learning models that sacrifice interpretability for predictive power, our architecture achieves state-of-the-art risk prediction through a meticulously designed shallow network that explicitly encodes domain knowledge, causal reasoning, and expert judgment as differentiable components. The network comprises 80 interpretable neurons distributed across 12 layers, including a gatekeeper layer that enforces five epistemological axioms (precision, causality, falsifiability, transparency, and completeness) as hard constraints before any propagation occurs. We demonstrate that this shallow architecture, despite its limited depth, exhibits deep learning characteristics through a non-linear residual attention mechanism and feedback loops, enabling it to learn complex, context-dependent risk patterns without becoming a "black box." The model produces fully decomposable risk scores – a deterministic weighted component and an expert adjustment component – with every adjustment traceable to named amplifiers (blast radius, propagation speed, structural nature, default exposure, exploitation pattern, and institutional criticality). We validate the architecture on 20 open-source projects spanning all OWASP Top 10:2025 categories and language risk classes, achieving prediction confidence scores between 0.79 and 0.97 — precision-recall-derived certainty estimates from bootstrap validation against confirmed CVE incidents in CISA KEV and NVD, not softmax-style output probabilities — and demonstrating that the architectural design itself – not a training algorithm – guarantees explainability by construction. This work addresses the widely held association between network depth and representational power [22, 23], showing that a shallow network with deep reasoning can outperform opaque models in high-stakes cybersecurity domains where interpretability is non-negotiable.

**Keywords:** Explainable AI, cybersecurity risk assessment, hybrid neural-symbolic systems, shallow neural networks, OWASP Top 10, CWE, causal reasoning.

# 1 Introduction

The deployment of machine learning in cybersecurity has been characterized by a persistent tension between predictive accuracy and interpretability [1]. Deep neural networks, while achieving remarkable performance in pattern recognition tasks, operate as opaque "black boxes" that obscure the reasoning behind their predictions [2]. This opacity is particularly problematic in cybersecurity, where decisions affect critical infrastructure, financial systems, and national security [3]. When a security analyst must decide whether to patch a vulnerability, allocate resources, or accept risk, the inability to understand *why* a model made a particular prediction can be as damaging as an incorrect prediction itself.

Existing approaches to cybersecurity risk scoring – whether purely statistical (e.g., CVSS scoring) or purely correlational (e.g., CVE counts) – suffer from fundamental limitations: they either lack predictive power or fail to provide actionable explanations for their outputs [7, 8]. The Common Vulnerability Scoring System (CVSS) provides transparency but is primarily descriptive, scoring vulnerabilities after discovery rather than predicting them [4]. Machine learning approaches that predict vulnerabilities offer superior forecasting but at the cost of explainability [9, 10]. This creates a dangerous gap: security teams must either trust opaque predictions without understanding their basis, or forego predictive power entirely.

The consequences of this gap are severe. High-profile breaches – Equifax (Apache Struts), Log4Shell (Apache Log4j 2), Heartbleed (OpenSSL), and countless others – demonstrate that the most damaging vulnerabilities are not those with the highest CVSS scores, but those with structural properties that make them inevitable and widespread. Predicting such vulnerabilities requires understanding *why* they occur, not merely correlating historical data. This necessitates a shift from observational to causal reasoning, and from opaque to transparent models.

The cybersecurity community urgently needs prediction models that are simultaneously causal (grounded in structural inevitability rather than historical correlation), explainable (with every output decomposable into named, meaningful components), falsifiable (specifying conditions under which predictions would be refuted), and complete (declaring all parameters influencing the prediction). No existing model satisfies all four criteria. This paper addresses this gap through a novel architectural design that builds all four criteria directly into the network structure itself.

Our primary contributions are theoretical (a hybrid neuro-Bayesian-symbolic residual attention shallow network architecture that encodes epistemological axioms as hard constraints), practical (a fully decomposable cybersecurity risk scoring system validated on 20 open-source projects), architectural (a 12-layer, 80-neuron network where every neuron has a named semantic function), methodological (a demonstration that a shallow network can achieve deep learning through explicit reasoning depth rather than layer depth), and empirical (comprehensive validation with full scoring decomposition and cross-validation against CISA KEV catalog and NVD data).

# 2 Related Work

## 2.1 Vulnerability Prediction Models

Existing vulnerability prediction approaches can be categorized into three broad classes. Statistical methods such as CVSS [4], CWE Top 25 [5], and OWASP Top 10 [6] provide descriptive vulnerability scoring but lack predictive power. They categorize known vulnerabilities without forecasting future threats.

Machine learning models including Random Forest, SVM, and neural networks have been applied to vulnerability prediction with varying success [7, 8]. Ghaffarian and Shahriari [8] survey ML-based vulnerability prediction, finding that while accuracy has improved, model interpretability has declined. These models typically operate on features extracted from source code (complexity metrics, code churn, developer activity) and predict whether a component contains vulnerabilities. They are predictive but fail to provide causal explanations.

Deep learning approaches including CNN-based vulnerability detection and RNN-based CVE prediction achieve superior performance on large datasets but provide limited explanations for their predictions [9, 10]. Li et al. [9] developed VulDeePecker, a deep learning system that detects vulnerabilities in source code using a bidirectional LSTM. While achieving high accuracy, the system functions as a black box – security analysts cannot understand why a particular code snippet was flagged as vulnerable.

## 2.2 Explainable AI in Cybersecurity

Explainable AI (XAI) for cybersecurity has received increasing attention. Approaches include LIME (Local Interpretable Model-agnostic Explanations) [11], which generates explanations for individual predictions by perturbing inputs and observing changes in output; SHAP (SHapley Additive exPlanations) [12], which uses Shapley values from game theory to attribute model output to features; feature attribution methods [13] that identify which input features most influence predictions; and rule extraction [14] that derives interpretable rules from trained models.

However, these methods provide *post-hoc* explanations that may not faithfully represent model reasoning [15]. A model might make a prediction for one reason but produce a different explanation through post-hoc methods. Our approach differs fundamentally: we build explainability into the architecture itself (*ante-hoc* explainability), ensuring that every prediction is transparent by design, not as an afterthought.

Table 1 compares post-hoc explanation methods with the NBS-RASN on eight dimensions central to trustworthy cybersecurity prediction. Where LIME and SHAP approximate model behaviour from outside, the NBS-RASN encodes its reasoning directly into the architecture: every explanation is an exact structural consequence, not a statistical approximation. The inference complexity of the NBS-RASN is $O(N) = O(80)$ — a fixed constant independent of dataset size — whereas LIME and SHAP scale with $O(s \cdot d)$, where $s$ is the number of perturbation samples and $d$ is the input dimensionality.

Table 1: Post-hoc explanation methods vs. ante-hoc explainability in NBS-RASN. $s$ = samples used to fit the local surrogate; $d$ = input dimensionality; $N$ = network size (NBS-RASN: $N = 80$).

| **Property** | **LIME [11]** | **SHAP [12]** | **NBS-RASN** |
|---|---|---|---|
| Explanation type | Post-hoc | Post-hoc | Ante-hoc |
| Fidelity guarantee | Approximate | Approximate | Exact (by construction) |
| Requires trained model | Yes | Yes | No (design-time) |
| Causal grounding | No | No | Yes |
| Falsifiability | No | No | Yes (Axiom A3) |
| Parameter transparency | No | No | Yes (Axiom A5) |
| Score decomposability | Partial | Partial | Complete |
| Inference complexity | $O(s \cdot d)$ | $O(s \cdot d)$ | $O(N) = O(80)$ |

## 2.3 Hybrid Neural-Symbolic Systems

Neuro-symbolic AI combines neural networks with symbolic reasoning [16]. Applications include neural program synthesis [17], knowledge graph reasoning [18], and causal inference with neural networks [19]. Our architecture extends this paradigm by encoding ontological axioms as hard constraints and employing causal reasoning as a first-class operation, not as a derived statistical property. The key innovation is that axioms are not learned – they are imposed as gatekeeper constraints that must be satisfied for any prediction to propagate.

## 2.4 Residual and Attention Networks

Residual Networks (ResNet) introduced skip connections to enable training of very deep networks [20]. Attention mechanisms, particularly the Transformer architecture [21], allow networks to focus on relevant input features dynamically. Our architecture leverages residual connections and a multi-head attention mechanism (the six expert amplifiers) within a shallow network. The residual connection allows the network to start with a baseline prediction and then adjust it through expert judgment, mirroring human decision-making. The attention heads are pre-defined and interpretable, unlike the learned attention heads in Transformers.

### 2.5 The Gap: Explainable Deep Learning on Shallow Networks

A critical observation emerges from the literature: deep learning has been defined by network depth [22, 23]. Shallow networks are considered incapable of learning complex patterns, while deep networks are powerful but opaque [22, 23]. The prevailing assumption is that depth is necessary for learning hierarchical representations. Our work addresses this tension directly, showing that a shallow network with high reasoning depth can achieve learning capabilities comparable to deep networks through explicit reasoning depth, residual attention, feedback loops, causal reasoning, and hybrid processing. The network learns because feedback from cross-validation updates expert adjustments, enabling improvement over time. For a representative application of deep learning to vulnerability prediction, see [24].

## 3 The Architecture

### 3.1 Foundational Definitions

The following three definitions precisely characterise the NBS-RASN as a typed graph, establish its gatekeeper constraint, and state its data-control separation principle. They are stated here, before the per-layer description, to provide a formal basis for the architecture narrative.

**Definition 1** (NBS-RASN). *A Neuro-Bayesian-Symbolic Residual Attention Shallow Network is a 12-layer graph* $G = (V, E, A, F)$ *where:*

- $V = \{v_0, v_1, \ldots, v_n\}$ *is a finite set of interpretable neurons (*$n = 80$*), each with a named semantic function*
- $E = \{(v_i, v_j)\}$ *is a set of weighted connections (195 total), including feed-forward connections, skip connections, and feedback loops*
- $A = \{A_1, A_2, \ldots, A_5\}$ *is a set of epistemological axioms enforced by a gatekeeper neuron*
- $F = \{f_0, f_1, \ldots, f_{11}\}$ *is a set of layer transformations, each with semantic interpretation*

**Definition 2** (Gatekeeper Constraint). *The gatekeeper neuron* $v_1_valid$ *is defined as:*

$$v_1_valid = \bigwedge_{i=1}^{5} A_i(v_0)$$

*where* $A_i(v_0) \in \{0, 1\}$ *indicates whether input vector* $v_0$ *satisfies axiom* $i$*. If* $v_1_valid = 0$*, propagation halts.*

**Definition 3** (Data-Control Separation). *The NBS-RASN employs data-control separation where data connections transport values (e.g., Layer 0 → Layer 2 direct connections), control signals gate propagation (e.g.,* $v_1_valid$*), and data and control are independent topologically but dependent functionally: data flows on direct connections but is gated by control signals.*

### 3.2 Overview: The Neuro-Bayesian-Symbolic Residual Attention Shallow Network

The network is organized into **12 layers** (0–11) with **80 interpretable neurons**. Each layer performs a specific reasoning step, and every neuron has a named semantic function that maps directly to cybersecurity reasoning. The network has 195 connections between neurons, including feed-forward connections, direct connections (Layer 0 → Layer 2) that bypass Layer 1 while remaining conditional on its gatekeeper signal, residual connections (Layer 5 → Layer 7, Layer 5 → Layer 10, Layer 7 → Layer 10), and a feedback loop (Layer 8 → Layer 7).

| Layer | Name | Neurons |
|---|---|---|
| 0 | Input Layer | 8 |
| 1 | Gatekeeper (Axiom Validation) | 6 |
| 2 | Causal Modeling | 11 |
| 3 | Language Vulnerability Encoding | 9 |
| 4 | Factor Extraction | 5 |
| 5 | Deterministic Weighting | 7 |
| 6 | Project Filtering | 5 |
| 7 | Expert Adjustment (Residual Attention) | 8 |
| 8 | Cross-Validation with Feedback | 6 |
| 9 | Representativeness Verification | 4 |
| 10 | Interactive Formatting | 5 |
| 11 | Output Layer | 6 |

### 3.3 Layer 0: Input Layer

The input layer receives 8 structured inputs describing a software project: 'N_lang' (primary programming language), 'N_surface' (attack surface reachability, continuous [1,10]), 'N_eco' (ecosystem complexity, continuous [1,10]), 'N_cve' (historical CVE rate, continuous [1,10]), 'N_sec' (security program gap, continuous [1,10]), 'N_struct' (structural properties as binary vector), 'N_incidents' (anchoring incidents as set of CVEs), and 'N_qualitative' (qualitative properties as categorical vector).

### 3.4 Layer 1: The Gatekeeper – Epistemological Axioms as Hard Constraints

Layer 1 is the architectural innovation that ensures every prediction satisfies five epistemological axioms. This is not a trainable layer; it is a hard constraint that must be satisfied before propagation continues. The gatekeeper mechanism embodies the principle that in high-stakes cybersecurity, not all inputs should produce outputs – some inputs should be rejected outright.

The five axioms are:

**A1: Vocabulary Precision** – Every susceptibility claim must resolve to at least one CWE identifier and one OWASP Top 10:2025 category. Claims without this resolution are invalid.

**A2: Causal Primacy** – The primary question is structural inevitability: what project properties make a vulnerability class physically achievable? Historical CVE counts are evidence, not cause.

**A3: Falsifiability** – Every prediction must state its falsification condition: which CWE classes are predicted, and what evidence would confirm or refute the prediction.

**A4: Deterministic-Expert Separation** – The final risk score is always the sum of a deterministic weighted-average component and a named, decomposed expert adjustment component. The two must always be reported separately.

**A5: Transparency of Hidden Parameters** – All parameters used in scoring must be declared in the hidden-parameters-manifest section and tagged with their step-origin.

The gatekeeper neuron is:

$$N1_valid = A1 \wedge A2 \wedge A3 \wedge A4 \wedge A5$$

**Theorem 4** (Gatekeeper Necessity)**.** *For any prediction to be epistemologically valid in a cybersecurity context, it must satisfy axioms A1–A5 simultaneously (an AND condition, not OR). Failure of any axiom renders the prediction invalid.*

*Proof.* By construction: A1 ensures semantic validity (mapability to standards); A2 ensures causal (not merely correlational) reasoning; A3 ensures scientific falsifiability; A4 ensures decomposable explainability; A5 ensures complete parameter transparency. The conjunction of all five is necessary and sufficient for epistemological validity under the OSCPM framework. Without any single axiom, the prediction is either semantically ungrounded (A1), merely correlational (A2), unfalsifiable (A3), opaque (A4), or incomplete (A5). Therefore, all five are necessary.

### 3.5 Layer 2: Causal Modeling

Layer 2 encodes **causal relationships** between project properties and vulnerability classes. Each neuron implements a cause-effect mapping derived from first principles, not from statistical correlation. The causal mappings are categorical (not probabilistic): if a project is written in C, it categorically enables memory corruption vulnerabilities. This is derived from the language's memory model.

The 11 causal neurons (N2.1–N2.11) correspond exactly to the 11 causal axioms defined in the OSCPM ontology. The OSCPM causal-chain template declares 7 top-level causal links; the C/C++ memory management chain is granularly expanded into three sub-neurons (OOB-Write / Buffer-Overflow, Stack/Heap-Overflow, UAF / OOB-Read), and JavaScript prototype pollution and Python/JavaScript dynamic evaluation are added as two further neurons, yielding 11. This one-to-one correspondence between neurons and causal axioms is what makes Layer 2 interpretable by construction: every activation can be traced to a named causal claim.

The activation function for each causal neuron is:

$$N2_i = F_input_i(v0) \wedge N1_valid$$

where $F_input_i(v0)$ is the value of the input neuron connected to $N2_i$. This ensures that causal neurons only activate when both the causal condition is satisfied AND the gatekeeper permits propagation.

### 3.6 Layer 3: Language Vulnerability Encoding

Layer 3 maps languages to risk scores based on their structural vulnerability model. Each language risk score is derived from first principles of the language's memory model and standard library idioms. For multi-language projects, a weighted average is computed with a bonus if the highest-risk language handles the primary attack surface.

| Language | Risk Score | Justification |
|---|---|---|
| C/C++ | 10 | Manual memory management; no bounds checking |
| PHP | 9 | String concatenation idioms; injection vulnerabilities |
| Java | 7 | Object serialization; expression language evaluation |
| JavaScript | 7 | Dynamic evaluation; npm supply chain exposure |
| Python | 6 | Pickle serialization; SSTI via template rendering |
| Go | 4 | Garbage-collected; runtime bounds-checking |
| Ruby | 6 | Mass-assignment; metaprogramming |

### 3.7 Layer 4: Factor Extraction

Layer 4 extracts the five risk factors as a normalized vector:

$$V = [lang_score, surface_score, eco_score, cve_score, sec_score]$$

Each factor has a detailed scale anchored to qualitative descriptions. The extraction process transforms project data into these five factors using the scale anchors as reference points, ensuring consistent scoring across projects.

### 3.8 Layer 5: Deterministic Weighting

Layer 5 computes the deterministic baseline score through a weighted sum:

$$det_score = \sum_{f \in \{lang, surface, eco, cve, sec\}} factor_value[f] \times weight[f]$$

The weights are derived from causal importance analysis: lang=0.25, surface=0.25, eco=0.20, cve=0.20, sec=0.10. The deterministic score serves as a disciplined baseline that prevents expert judgment from dominating arbitrarily.

### 3.9 Layer 6: Project Filtering

Layer 6 validates project representativeness using criteria derived from the OSCPM ontology. The filtering neuron is:

$$N6_valid = N6_1 \wedge N6_2 \wedge N6_3 \wedge N6_4$$

where N6_1 validates OWASP coverage, N6_2 validates language coverage, N6_3 validates incident anchoring, and N6_4 validates prediction confidence. This ensures that only projects meeting all mandatory criteria proceed to scoring.

### 3.10 Layer 7: Expert Adjustment (Residual Attention)

Layer 7 implements a **multi-head attention mechanism** where each "head" is a named amplifier that adjusts the deterministic score based on qualitative factors:

$$expert_adjustment = \sum_{j=1}^{6} amplifier_contribution[j]$$

$$final_score_raw = det_score + expert_adjustment$$

The six amplifiers are:

| Amplifier | Contribution Range | Semantic Meaning |
|---|---|---|
| Blast Radius | 0 – 0.60 | Number of downstream projects/organizations affected |
| Propagation Speed | 0 – 0.30 | Speed of exploit propagation to victims |
| Structural Nature | 0 – 0.60 | Whether vulnerability is architectural vs. incidental |
| Default Exposure | 0 – 0.70 | Default installation security posture |
| Exploitation Pattern | 0 – 0.40 | Exploitation campaign scale |
| Institutional Criticality | 0 – 0.55 | Importance to critical infrastructure |

### 3.11 Layer 8: Cross-Validation with Feedback

Layer 8 validates predictions against empirical evidence and provides **feedback** to Layer 7 through a learning mechanism. The validation metrics are precision ($\geq 0.80$), recall ($\geq 0.70$), Spearman $\rho$ ($\geq 0.70$), temporal holdout ($\leq 0.20$ rank inversion), and expert agreement ICC ($\geq 0.75$). The feedback loop enables the network to **learn** from empirical validation, updating the expert adjustment without retraining the entire architecture.

### 3.12 Layer 9: Representativeness Verification

Layer 9 verifies coverage across the intended design space: all 10 OWASP categories with at least two primary exemplars, all language risk classes (C, PHP, Java, JavaScript, Python, Go, Ruby), and diverse deployment contexts. It also identifies coverage gaps (A09 weakest, Ruby-primary project absent, Rust absent).

### 3.13 Layer 10: Interactive Formatting

Layer 10 formats the output for operational use: risk level classification (CRITICAL / HIGH / MEDIUM / MODERATE), score decomposition, amplifier visualization, attack scenario links, and interactive filters.

### 3.14 Layer 11: Output Layer

The output layer produces: final numeric score, risk level classification, predicted CWE classes, OWASP categories, actionable recommendations, and attack scenario link.

# 4 Theoretical Foundations

## 4.1 Structural Properties and Direct Connections

Definitions 1–3 (Section 3.1) establish the graph structure, gatekeeper constraint, and data-control separation. The following theorem establishes the key structural property governing direct connections between Layer 0 and Layer 2.

**Theorem 5** (Existence of Direct Connections with Conditional Propagation)**.** *The NBS-RASN contains direct connections from Layer 0 to Layer 2 that bypass Layer 1, yet propagation through these connections is conditionally dependent on the gatekeeper output.*

*Proof.* By construction, connections $(v_0, v_{2j})$ exist for causal mapping neurons $j = 1..11$. However, each neuron $v_{2j}$ has activation function:

$$\text{act}(v_{2j}) = F_input(v_{0j}) \wedge v_1_valid$$

where $F_input(v_{0j})$ is the value of the input neuron connected to $v_{2j}$. The gatekeeper signal $v_1_valid$ thus controls the AND gate at each causal neuron's activation. This is a data-control separation architecture where data connections exist topologically but propagation is gated functionally.

## 4.2 The Learning Mechanism: Deep Learning Without Deep Network

**Definition 6** (Deep Learning Characteristics)**.** *A network exhibits* deep learning characteristics *if it demonstrates at least four of the following properties:*

1. ***Non-linear function approximation**: The mapping from inputs to outputs is non-linear and context-dependent, enabling representation of complex, hierarchical decision boundaries that linear models cannot capture.*

2. ***Residual learning** [20]: The network learns a residual correction to a baseline rather than a direct output mapping, enabling gradient-free analogs of residual optimisation and simplifying the learning objective.*

3. ***Multi-head attention** [21]: The network employs multiple parallel attention mechanisms, each focusing on a qualitatively distinct dimension of the input space, allowing simultaneous modelling of different aspects of the input.*

4. ***Feedback-based weight update**: Network parameters are adjusted based on empirical validation error, enabling iterative improvement without requiring full backpropagation.*

The NBS-RASN exhibits deep learning characteristics despite being a shallow network through four mechanisms:

**1. Non-linear Residual Attention.** The expert adjustment layer provides a non-linear correction to the deterministic baseline:

$$final_score = det_score + G(qualitative_factors, empirical_evidence)$$

where $G$ is a piecewise-linear function implemented by the six amplifiers. This is analogous to residual learning in deep networks where residual mapping is easier to learn than direct mapping.

**2. Feedback-Based Learning.** The feedback loop from Layer 8 to Layer 7 enables empirical validation to update the expert adjustment:

$$amplifier_contribution_{t+1} = amplifier_contribution_t + \eta \cdot validation_error$$

where $\eta$ is a learning rate calibrated from cross-validation results. This is a form of gradient-free learning.

**3. Attention Mechanism.** The six amplifiers implement a multi-head attention mechanism:

$$expert_adjustment = \sum_{h=1}^{6} attention_head_h(qualitative_inputs)$$

where each head focuses on a specific qualitative dimension. Unlike Transformer attention heads that learn weights from data, these heads are pre-defined with fixed semantics.

**4. Sequential Reasoning Depth.** The network performs 11 distinct reasoning steps in sequence, each corresponding to a stage of expert reasoning.

**Definition 7** (Reasoning Depth vs. Layer Depth)**.** *Reasoning depth is the number of distinct inferential steps performed in sequence, independent of the number of neurons or layers. The NBS-RASN has reasoning depth 11, while layer depth is 12 (including input and output layers).*

### 4.3 The Role of Axioms as Hard Constraints

The epistemological axioms A1–A5 (Section 3.1) function as hard constraints enforced before any prediction propagates. Two formal guarantees follow directly from this design; their proofs are in Appendix A:

- **Axiom-Induced Falsifiability** (Theorem 10): Every prediction produced by an NBS-RASN is falsifiable by construction (via Axiom A3).
- **Complete Transparency** (Theorem 11): Every parameter influencing NBS-RASN scoring is declared in a manifest (via Axiom A5).

Table 2 compares the NBS-RASN against conventional neural architectures on dimensions relevant to explainable cybersecurity assessment. The NBS-RASN meets all four criteria of Definition 6 within 12 layers — the same depth as the base Transformer encoder stack — confirming that shallow layer count does not preclude deep learning characteristics.

Table 2: Comparison of neural architectures on dimensions relevant to explainable AI.

| Feature | MLP | ResNet | Transformer | NBS-RASN |
|---|---|---|---|---|
| Number of layers | 3–10 | 50–152 | 12–96 | 12 (shallow) |
| Depth | Varies | Deep | Deep | Shallow (12 layers) |
| Interpretability | Low | Low | Medium | Complete |
| Causal Reasoning | No | No | No | Yes |
| Axiomatic Constraints | No | No | No | Yes |
| Attention | No | No | Yes (learned) | Yes (interpretable) |
| Residual Connections | No | Yes | Yes | Yes |
| Neuron Interpretability | No | No | No | Yes (named) |
| Prediction Decomposition | No | No | No | Yes |
| Falsifiability Guarantee | No | No | No | Yes |
| Parameter Transparency | No | No | No | Yes |

## 5 Formal Specification: Vocabulary, Language, and Theory

The OSCPM v1.0 ontology defines a structured knowledge base whose computational realisation is the NBS-RASN. We now derive the cascading formal construct—Vocabulary $\Sigma$, Formal Language $\mathcal{L}(\Sigma)$, and Formal Theory $\mathcal{T}(\Sigma, \mathcal{L})$—that precisely characterises the input-output relation at every Step of the ontology and the corresponding Layer of the network. The triple $(\Sigma, \mathcal{L}, \mathcal{T})$ is a *typed many-sorted first-order theory*: $\Sigma$ provides the alphabet, $\mathcal{L}(\Sigma)$ provides the grammar, and $\mathcal{T}$ provides the axioms and inference rules.

## 5.1 Vocabulary $\Sigma$

### 5.1.1 Sorts (Base Types)

$\Sigma$ is typed over three primitive numeric sorts and nine enumeration sorts:

| Sort symbol | Domain | Remark |
|---|---|---|
| $\mathbb{B}$ | $\{0,1\}$ | Boolean (gatekeeper outputs, validation flags) |
| $\mathbb{I}$ | $[0,10] \subset \mathbb{R}$ | Score domain (all risk scores) |
| $\mathbb{W}$ | $(0,1) \subset \mathbb{R}$ | Weight domain |
| $\mathcal{S}_\Lambda$ | $\{\texttt{C_Cpp}, \texttt{PHP}, \texttt{Java}, \texttt{JS}, \texttt{Python}, \texttt{Go}, \texttt{Ruby}\}$ | Language identifiers (Layer 3) |
| $\mathcal{S}_C \subset \mathbb{N}$ | CWE Top 25:2025 identifiers | CWE numbers (Layers $2, 11$) |
| $\mathcal{S}_O$ | $\{\texttt{A01}, \ldots, \texttt{A10}\}$ | OWASP Top 10:2025 categories |
| $\mathcal{S}_B$ | $\{\texttt{NONE}, \texttt{LOW}, \texttt{MEDIUM}, \texttt{HIGH}, \texttt{CRITICAL}\}$ | Blast-radius levels (Layer 7) |
| $\mathcal{S}_P$ | $\{\texttt{SLOW}, \texttt{MODERATE}, \texttt{FAST}, \texttt{INSTANT}\}$ | Propagation-speed levels |
| $\mathcal{S}_S$ | $\{\texttt{INCIDENTAL}, \texttt{MIXED}, \texttt{STRUCTURAL}, \texttt{FOUNDATIONAL}\}$ | Structural-nature levels |
| $\mathcal{S}_D$ | $\{\texttt{NONE}, \texttt{PARTIAL}, \texttt{FULL}\}$ | Default-exposure levels |
| $\mathcal{S}_E$ | $\{\texttt{INDIVIDUAL}, \texttt{TARGETED}, \texttt{MASS}\}$ | Exploitation-pattern levels |
| $\mathcal{S}_I$ | $\{\texttt{LOW}, \texttt{MEDIUM}, \texttt{HIGH}, \texttt{CRITICAL}\}$ | Institutional-criticality levels |
| $\mathcal{S}_R$ | $\{\texttt{MODERATE}, \texttt{MEDIUM}, \texttt{HIGH}, \texttt{CRITICAL}\}$ | Output risk-level classes |
| $\mathcal{S}_\Pi$ | $\{\text{npm}, \ldots\}$, $\lvert\mathcal{S}_\Pi\rvert = 20$ | Project identifiers (corpus) |

### 5.1.2 Constants

**Factor weights.** $w_{\text{lang}} = 0.25,\ w_{\text{surf}} = 0.25,\ w_{\text{eco}} = 0.20,\ w_{\text{cve}} = 0.20,\ w_{\text{sec}} = 0.10;\ \sum_k w_k = 1.0.$

**Language risk scores.** $\varrho : \mathcal{S}_\Lambda \to \mathbb{I}$: $\varrho(\texttt{C_Cpp}) = 10$, $\varrho(\texttt{PHP}) = 9$, $\varrho(\texttt{Java}) = \varrho(\texttt{JS}) = 7$, $\varrho(\texttt{Python}) = \varrho(\texttt{Ruby}) = 6$, $\varrho(\texttt{Go}) = 4$.

**Amplifier contribution functions** (normative level-table values):

$$\alpha_{\text{B}} : \mathcal{S}_B \to \{0, 0.10, 0.25, 0.50, 0.60\}, \quad \alpha_{\text{P}} : \mathcal{S}_P \to \{0, 0.10, 0.20, 0.30\}, \quad \alpha_{\text{S}} : \mathcal{S}_S \to \{0, 0.20, 0.40, 0.60\}$$

$$\alpha_{\text{D}} : \mathcal{S}_D \to \{0, 0.20, 0.70\}, \quad \alpha_{\text{E}} : \mathcal{S}_E \to \{0, 0.20, 0.40\}, \quad \alpha_{\text{I}} : \mathcal{S}_I \to \{0, 0.10, 0.30, 0.55\}$$

defined point-wise by the level tables in Section 3.7. Per-project contributions are expert-determined marginal adjustments; the level functions provide calibrated defaults bounded by the range maxima shown above.

**Validation thresholds.** $\tau_{\text{prec}} = 0.80$, $\tau_{\text{rec}} = 0.70$, $\tau_\rho = 0.70$, $\tau_{\text{inv}} = 0.20$, $\tau_{\text{ICC}} = 0.75$, $\tau_{\text{conf}} = 0.75$.

### 5.1.3 Function Symbols

| Symbol | Type signature | Semantic role |
|---|---|---|
| det | $\mathbb{I}^5 \to \mathbb{I}$ | Deterministic weighted score (Layer 5) |
| adj | $\mathcal{S}_B \times \mathcal{S}_P \times \mathcal{S}_S \times \mathcal{S}_D \times \mathcal{S}_E \times \mathcal{S}_I \to \mathbb{R}$ | Expert adjustment (Layer 7) |
| final | $\mathbb{I} \times \mathbb{R} \to \mathbb{I}$ | Clamped final score (Layer 7) |
| gate | $\mathcal{S}_\Pi \to \mathbb{B}$ | Gatekeeper output (Layer 1) |
| $\varrho$ | $\mathcal{S}_\Lambda \to \mathbb{I}$ | Language risk score (Layer 3) |
| causal | $\mathcal{S}_\Lambda \to \mathcal{P}(\mathcal{S}_C)$ | Language-enabled CWE classes (Layer 2) |
| prevents | $\mathcal{S}_\Lambda \to \mathcal{P}(\mathcal{S}_C)$ | Language-impossible CWE classes (Layer 2) |
| classify | $\mathbb{I} \to \mathcal{S}_R$ | Risk-level classification (Layer 10) |
| clamp | $\mathbb{R} \times \mathbb{R} \times \mathbb{R} \to \mathbb{R}$ | $\text{clamp}(x, a, b) = \max(a, \min(b, x))$ |

### 5.1.4 Predicate Symbols

| Symbol | Arity / Types | Meaning |
|---|---|---|
| $\text{valid}(p)$ | $\mathcal{S}_\Pi$ | $p$ satisfies all five axioms A1–A5 |
| $\text{satisfies}(p, A_i)$ | $\mathcal{S}_\Pi \times \{A_1, \ldots, A_5\}$ | $p$ satisfies axiom $A_i$ |
| $\text{enables}(\ell, c)$ | $\mathcal{S}_\Lambda \times \mathcal{S}_C$ | Language $\ell$ structurally enables CWE $c$ |
| $\text{causal_link}(\pi, c)$ | $\text{Prop} \times \mathcal{S}_C$ | Arch. property $\pi$ causally enables CWE $c$ |
| $\text{predicted}(p, c)$ | $\mathcal{S}_\Pi \times \mathcal{S}_C$ | CWE $c$ is predicted for project $p$ |
| $\text{confirmed}(p, c)$ | $\mathcal{S}_\Pi \times \mathcal{S}_C$ | CWE $c$ is empirically confirmed for $p$ |
| $\text{covers}(p, o)$ | $\mathcal{S}_\Pi \times \mathcal{S}_O$ | $p$ covers OWASP category $o$ |
| $\text{propagates}(p)$ | $\mathcal{S}_\Pi$ | $p$ passes the gatekeeper and continues |

## 5.2 Formal Language $\mathcal{L}(\Sigma)$

$\mathcal{L}(\Sigma)$ is the smallest set of well-formed terms and formulas closed under the rules below.

### 5.2.1 Term Grammar

For project $p \in \mathcal{S}_\Pi$ with factor scores $f_k(p) \in \mathbb{I}$ and amplifier levels $\ell_j(p)$ in their respective sorts:

$$
\begin{aligned}
t_{\text{det}}(p) &::= \textstyle\sum_{k\in K} w_k \cdot f_k(p), \quad K = \{\text{lang}, \text{surf}, \text{eco}, \text{cve}, \text{sec}\} && (1)\\
t_{\text{adj}}(p) &::= \alpha_{\text{B}}(\ell_B) + \alpha_{\text{P}}(\ell_P) + \alpha_{\text{S}}(\ell_S) + \alpha_{\text{D}}(\ell_D) + \alpha_{\text{E}}(\ell_E) + \alpha_{\text{I}}(\ell_I) && (2)\\
t_{\text{final}}(p) &::= \text{clamp}(t_{\text{det}}(p) + t_{\text{adj}}(p),\ 0,\ 10) && (3)\\
t_{\text{lang}}(p) &::= \varrho(n_{\text{lang}}(p)) \ \mid\ \textstyle\sum_i \text{frac}_i \cdot \varrho(\ell_i) + \delta_{\text{bonus}} && (4)
\end{aligned}
$$

where $n_{\text{lang}}(p) \in \mathcal{S}_\Lambda$ is the primary language identifier and $\delta_{\text{bonus}} \in \{0,1\}$ is a multi-language surface bonus.

### 5.2.2 Formula Schemata

Atomic formulas are predicate applications to well-typed terms. Compound formulas are built by $\{\wedge, \vee, \neg, \rightarrow, \forall x{:}s., \exists x{:}s.\}$ over any sort $s \in \Sigma$. The five epistemological axiom formulas and their conjunction (gatekeeper) are:

$$
\begin{aligned}
\Phi_{A1}(p) &\equiv \exists c \in \mathcal{S}_C.\, \text{predicted}(p,c) \ \wedge\ \exists o \in \mathcal{S}_O.\, \text{covers}(p,o) && (5)\\
\Phi_{A2}(p) &\equiv \text{causal_basis}(p) \ \wedge\ \neg\text{purely_statistical}(p) && (6)\\
\Phi_{A3}(p) &\equiv \exists \mathcal{F}(p) \subseteq \mathcal{S}_C.\ \mathcal{F}(p) \neq \emptyset \ \wedge\ \text{stated}(\mathcal{F}(p)) && (7)\\
\Phi_{A4}(p) &\equiv t_{\text{final}}(p) = t_{\text{det}}(p) + t_{\text{adj}}(p) \ \wedge\ \text{reported}(t_{\text{det}}) \ \wedge\ \text{reported}(t_{\text{adj}}) && (8)\\
\Phi_{A5}(p) &\equiv \forall \pi \in \text{Params}(p).\ \text{declared}(\pi, \textit{Manifest}) && (9)\\
\Phi_{\text{gate}}(p) &\equiv \Phi_{A1}(p) \wedge \Phi_{A2}(p) \wedge \Phi_{A3}(p) \wedge \Phi_{A4}(p) \wedge \Phi_{A5}(p) && (10)
\end{aligned}
$$

### 5.2.3 Well-Formedness Conditions

A term $t$ of declared sort $s$ is *well-formed* iff all sub-term applications match their declared signatures in $\Sigma$. A formula $\phi$ is *well-formed* iff it is a well-typed Boolean expression over well-formed terms with no sort-mismatch. $\mathcal{L}(\Sigma)$ is the set of all well-formed formulas over $\Sigma$.

## 5.3 Formal Theory $\mathcal{T}(\Sigma, \mathcal{L})$

$\mathcal{T}$ is the deductive closure of axiom set Ax under classical first-order logic. We organise Ax by layer, establishing the **Layer-Step Correspondence**: Layer $k$ of the NBS-RASN is the computational realisation of Step $k$ (with Step 0 = input acquisition) of the OSCPM v1.0 ontology. Every axiom group $\{\text{Ax}k.i\}_i$ corresponds to a distinct named neuron or neuron group within Layer $k$. We list each group together with its input-output type: $I_k \xrightarrow{F_k} O_k$.

**Layer 0 — Input Layer (Ontology Step 0)**

*Input:* external project data. *Output:* $O_0(p) = (n_{\text{lang}}, f_{\text{surf}}, f_{\text{eco}}, f_{\text{cve}}, f_{\text{sec}}, \mathbf{b}_{\text{struct}}, \mathcal{N}_{\text{inc}}, \mathbf{q}) \in \mathcal{S}_\Lambda \times \mathbb{I}^4 \times \mathbb{B}^* \times \mathcal{P}(\mathcal{S}_C) \times \text{CatVec}$.

**Ax0.1** $\forall p.\ f_k(p) \in [1, 10]$ for all $k \in K$.

**Ax0.2** $\forall p.\ \ell_B(p) \in \mathcal{S}_B,\ \ell_P(p) \in \mathcal{S}_P,\ \ell_S(p) \in \mathcal{S}_S,\ \ell_D(p) \in \mathcal{S}_D,\ \ell_E(p) \in \mathcal{S}_E,\ \ell_I(p) \in \mathcal{S}_I$.

**Ax0.3** $\forall p.\ n_{\text{lang}}(p) \in \mathcal{S}_\Lambda \ \wedge\ \mathcal{N}_{\text{inc}}(p) \subseteq \mathcal{S}_C$.

**Layer 1 — Gatekeeper (Ontology Steps 1–2)**

*Input:* $O_0(p)$. *Output:* $O_1(p) = \text{gate}(p) \in \mathbb{B}$.

**Ax1.1** (A1) $\forall p.\ \text{valid}(p) \rightarrow \Phi_{A1}(p)$.

**Ax1.2** (A2) $\forall p.\ \text{valid}(p) \rightarrow \Phi_{A2}(p)$.

**Ax1.3** (A3) $\forall p.\ \text{valid}(p) \to \Phi_{A3}(p)$.

**Ax1.4** (A4) $\forall p.\ \text{valid}(p) \to \Phi_{A4}(p)$.

**Ax1.5** (A5) $\forall p.\ \text{valid}(p) \to \Phi_{A5}(p)$.

**Ax1.6** (Gatekeeper AND-gate) $\forall p.\ \text{propagates}(p) \leftrightarrow \Phi_{\text{gate}}(p)$.

**Layer 2 — Causal Modeling (Ontology Steps 2–3)**

*Input:* $O_0(p) \times O_1(p)$. *Output:* $O_2(p) = \text{causal_CWE}(p) \subseteq \mathcal{S}_C$.
Activation gate: $\text{act}(\text{N2}_i, p) = F_{\text{input}_i}(O_0) \wedge O_1(p)$ for all 11 neurons.
The 11 axioms map bijectively to the 11 causal neurons N2.1–N2.11. Ax2.1–Ax2.3 expand the C/C++ memory chain into three sub-neurons; Ax2.4 handles PHP injection idioms; Ax2.5–Ax2.6 handle Java serialisation and expression language evaluation; Ax2.7–Ax2.9 handle three architectural patterns (large ecosystem, default no-auth, complex RBAC); Ax2.10–Ax2.11 handle JavaScript prototype pollution and Python/JS dynamic evaluation.

**Ax2.1** $\text{enables}(\texttt{C_Cpp}, 787) \wedge \text{enables}(\texttt{C_Cpp}, 120)$. *(OOB-Write, Buffer Overflow)*

**Ax2.2** $\text{enables}(\texttt{C_Cpp}, 121) \wedge \text{enables}(\texttt{C_Cpp}, 122)$. *(Stack/Heap Overflow)*

**Ax2.3** $\text{enables}(\texttt{C_Cpp}, 416) \wedge \text{enables}(\texttt{C_Cpp}, 125)$. *(UAF, OOB-Read)*

**Ax2.4** $\text{enables}(\texttt{PHP}, 89) \wedge \text{enables}(\texttt{PHP}, 79) \wedge \text{enables}(\texttt{PHP}, 352)$. *(SQL/XSS/CSRF)*

**Ax2.5** $\text{enables}(\texttt{Java}, 502) \wedge \text{prevents}(\texttt{Java}, 787) \wedge \text{prevents}(\texttt{Java}, 416)$.

**Ax2.6** $\text{enables}(\texttt{Java}, 94)$. *(EL/OGNL/SpEL evaluation)*

**Ax2.7** $\text{causal_link}(\text{large_ecosystem}, 1395)$. *(A03 supply-chain)*

**Ax2.8** $\text{causal_link}(\text{default_no_auth}, 306)$. *(A02/A07 missing auth)*

**Ax2.9** $\text{causal_link}(\text{complex_RBAC}, 862) \wedge \text{causal_link}(\text{complex_RBAC}, 284)$.

**Ax2.10** $\text{enables}(\texttt{JS}, 1321) \wedge \text{enables}(\texttt{JS}, 94)$. *(prototype pollution, eval)*

**Ax2.11** $\text{enables}(\texttt{Python}, 94) \wedge \text{causal_link}(\text{dynamic_eval}, 94)$. *(SSTI, exec)*

**Layer 3 — Language Vulnerability Encoding (Ontology Step 3)**

*Input:* $n_{\text{lang}}(p) \in \mathcal{S}_\Lambda$. *Output:* $O_3(p) = f_{\text{lang}}(p) \in \mathbb{I}$.
(9 neurons: Ax3.1–Ax3.7 fix the seven language risk constants; Ax3.8–Ax3.9 are single- and multi-language application rules.)

**Ax3.1–Ax3.7** $\varrho(\texttt{C_Cpp}) = 10$; $\varrho(\texttt{PHP}) = 9$; $\varrho(\texttt{Java}) = 7$; $\varrho(\texttt{JS}) = 7$; $\varrho(\texttt{Python}) = 6$; $\varrho(\texttt{Go}) = 4$; $\varrho(\texttt{Ruby}) = 6$.

**Ax3.8** $|\text{Lang}(p)| = 1 \ \to\ f_{\text{lang}}(p) = \varrho(n_{\text{lang}}(p))$.

**Ax3.9** $|\text{Lang}(p)| > 1 \ \to\ f_{\text{lang}}(p) = (\sum_i \text{frac}_i \cdot \varrho(\ell_i)) + \delta_{\text{bonus}},\ \ \delta_{\text{bonus}} \in \{0, 1\}$.

**Layer 4 — Factor Extraction (Ontology Step 4)**

*Input:* $O_0(p) \cup \{O_3(p)\}$. *Output:* $O_4(p) = (f_{\text{lang}}, f_{\text{surf}}, f_{\text{eco}}, f_{\text{cve}}, f_{\text{sec}}) \in \mathbb{I}^5$.

**Ax4.1–Ax4.5** $\forall p.\ f_k(p) \in [1, 10]$ for each $k \in \{\text{lang}, \text{surf}, \text{eco}, \text{cve}, \text{sec}\}$.

**Layer 5 — Deterministic Weighting (Ontology Step 5)**

*Input:* $O_4(p) \in \mathbb{I}^5$. *Output:* $O_5(p) = \mathrm{det}(p) \in \mathbb{I}$.
(7 neurons: Ax5.1–Ax5.5 fix the five weights; Ax5.6 enforces normalisation; Ax5.7 computes the weighted sum.)

**Ax5.1–Ax5.5** $w_{\mathrm{lang}} = 0.25$; $w_{\mathrm{surf}} = 0.25$; $w_{\mathrm{eco}} = 0.20$; $w_{\mathrm{cve}} = 0.20$; $w_{\mathrm{sec}} = 0.10$.

**Ax5.6** $w_{\mathrm{lang}} + w_{\mathrm{surf}} + w_{\mathrm{eco}} + w_{\mathrm{cve}} + w_{\mathrm{sec}} = 1$.

**Ax5.7** $\mathrm{det}(p) = \sum_{k \in K} w_k \cdot f_k(p)$.

**Layer 6 — Project Filtering (Ontology Step 6)**

*Input:* $O_0(p) \cup O_3(p) \cup O_4(p)$. *Output:* $O_6(p) = \mathrm{gate}_6(p) \in \mathbb{B}$.
(5 neurons: Ax6.1–Ax6.4 encode the four mandatory selection criteria SC1–SC4; Ax6.5 is the AND-gate neuron.)

**Ax6.1** (SC1) $\forall p \in \mathrm{corpus}.\ \exists o \in \mathcal{S}_O.\ \mathrm{covers}(p, o)$.

**Ax6.2** (SC2) $\forall \ell \in \mathcal{S}_\Lambda.\ \exists p \in \mathrm{corpus}.\ \mathrm{lang_class}(p) = \ell$.

**Ax6.3** (SC3) $\forall p \in \mathrm{corpus}.\ \exists c \in \mathcal{S}_C.\ \mathrm{confirmed}(p, c)$.

**Ax6.4** (SC4) $\forall p \in \mathrm{corpus}.\ \mathrm{confidence}(p) \geq \tau_{\mathrm{conf}} = 0.75$.

**Ax6.5** $\mathrm{gate}_6(p) = \mathrm{Ax6.1}(p) \wedge \mathrm{Ax6.2}(p) \wedge \mathrm{Ax6.3}(p) \wedge \mathrm{Ax6.4}(p)$.

**Layer 7 — Expert Adjustment / Residual Attention (Ontology Step 7)**

*Input:* $O_5(p) \times \mathcal{N}_{\mathrm{qual}}(p)$. *Output:* $O_7(p) = \mathrm{final}(p) \in \mathbb{I}$.
*Residual connections:* $O_5$ flows directly to Layer 7, implementing the residual term $t_{\mathrm{det}}$ (and additionally to Layers 10 and 11).
(8 neurons: Ax7.1–Ax7.6 define the six amplifier contribution functions; Ax7.7 sums the contributions; Ax7.8 adds the residual and clamps.)

**Ax7.1** $\alpha_{\mathrm{B}}$: NONE$\mapsto$0, LOW$\mapsto$0.10, MEDIUM$\mapsto$0.25, HIGH$\mapsto$0.50, CRITICAL$\mapsto$0.60.

**Ax7.2** $\alpha_{\mathrm{P}}$: SLOW$\mapsto$0, MODERATE$\mapsto$0.10, FAST$\mapsto$0.20, INSTANT$\mapsto$0.30.

**Ax7.3** $\alpha_{\mathrm{S}}$: INCIDENTAL$\mapsto$0, MIXED$\mapsto$0.20, STRUCTURAL$\mapsto$0.40, FOUNDATIONAL$\mapsto$0.60.

**Ax7.4** $\alpha_{\mathrm{D}}$: NONE$\mapsto$0, PARTIAL$\mapsto$0.20, FULL$\mapsto$0.70.

**Ax7.5** $\alpha_{\mathrm{E}}$: INDIVIDUAL$\mapsto$0, TARGETED$\mapsto$0.20, MASS$\mapsto$0.40.

**Ax7.6** $\alpha_{\mathrm{I}}$: LOW$\mapsto$0, MEDIUM$\mapsto$0.10, HIGH$\mapsto$0.30, CRITICAL$\mapsto$0.55.

**Ax7.7** $\mathrm{adj}(p) = \alpha_{\mathrm{B}}(\ell_B) + \alpha_{\mathrm{P}}(\ell_P) + \alpha_{\mathrm{S}}(\ell_S) + \alpha_{\mathrm{D}}(\ell_D) + \alpha_{\mathrm{E}}(\ell_E) + \alpha_{\mathrm{I}}(\ell_I)$.

**Ax7.8** $\mathrm{final}(p) = \mathrm{clamp}(\mathrm{det}(p) + \mathrm{adj}(p),\ 0,\ 10)$.

**Layer 8 — Cross-Validation with Feedback (Ontology Step 8)**

*Input:* $O_7(p) \times \mathcal{P}(\mathcal{S}_C)_{\mathrm{obs}}$. *Output:* $O_8(p) = (\mathbf{v}, \eta_{\mathrm{fb}}) \in \mathbb{B}^5 \times \mathbb{R}$.
(6 neurons: Ax8.1–Ax8.5 encode the five cross-validation methods CVM1–CVM5; Ax8.6 is the feedback-trigger neuron that implements the learning loop back to Layer 7.)

**Ax8.1** (CVM1) $\forall p.\ \dfrac{|\mathrm{predicted}(p) \cap \mathrm{confirmed}(p)|}{|\mathrm{predicted}(p)|} \geq \tau_{\mathrm{prec}}$.

**Ax8.2** (CVM2) $\rho_{\mathrm{Sp}}(\mathrm{rank}(O_7),\ \mathrm{rank}(\mathrm{CVE_freq})) \geq \tau_\rho$.

**Ax8.3** (CVM3) $\text{rank_inversion}(\text{predicted, holdout}) \le \tau_{\text{inv}}$.

**Ax8.4** (CVM4) $\text{ICC}(O_7, \text{expert_panel}) \ge \tau_{\text{ICC}}$.

**Ax8.5** (CVM5) $\forall p.\ |\text{adj}(p) - \sum_j \alpha_j(\ell_j(p))| \le \varepsilon_{\text{rnd}}$.

**Ax8.6** (Feedback) $\neg\text{Ax8.1}(p) \vee \cdots \vee \neg\text{Ax8.4}(p) \ \to\ \text{update}(\ell_j(p))$.

### Layer 9 — Representativeness Verification (Ontology Step 9)

*Input:* $\{O_6(p)\}_{p \in \text{corpus}}$. *Output:* $O_9 = (\text{repr_ok} \in \mathbb{B},\ \mathcal{G} \subseteq \mathcal{S}_O \cup \mathcal{S}_\Lambda)$.

**Ax9.1** $\forall o \in \mathcal{S}_O.\ |\{p \in \text{corpus} \mid \text{covers}(p, o)\}| \ge 2$.

**Ax9.2** $\forall \ell \in \mathcal{S}_\Lambda.\ \exists p \in \text{corpus}.\ \text{lang_class}(p) = \ell$.

**Ax9.3** $|\text{deploy_types}(\text{corpus})| \ge 5$.

**Ax9.4** $\exists \mathcal{G} \subseteq \mathcal{S}_O \cup \mathcal{S}_\Lambda.\ \mathcal{G} \ne \emptyset \wedge \text{documented_gap}(\mathcal{G})$.

### Layer 10 — Interactive Formatting (Ontology Step 10)

*Input:* $O_7(p) \times O_2(p)$. *Output:* $O_{10}(p) \in \mathbb{I} \times \mathcal{S}_R \times \mathcal{P}(\mathcal{S}_C) \times \mathcal{P}(\mathcal{S}_O) \times \text{AmpDecomp}$.

**Ax10.1–Ax10.4** Risk-level thresholds defining $\text{classify} : \mathbb{I} \to \mathcal{S}_R$: CRITICAL if $s \ge 9.0$; HIGH if $8.0 \le s < 9.0$; MEDIUM if $7.0 \le s < 8.0$; MODERATE if $s < 7.0$.

**Ax10.5** $O_{10}(p) = (\text{final}(p), \text{classify}(\text{final}(p)), O_2(p), \{o \mid \text{covers}(p, o)\}, \text{adj_decomp}(p))$.

### Layer 11 — Output Layer (Ontology Step 11)

*Input:* $O_{10}(p)$. *Output:* $O_{11}(p) \in \mathbb{I} \times \mathcal{S}_R \times \mathcal{P}(\mathcal{S}_C) \times \mathcal{P}(\mathcal{S}_O) \times \text{Recs} \times \text{Scenarios}$.

**Ax11.1–Ax11.6** The six output neurons produce the sextuple $O_{11}(p) = (\text{final}(p),\ \text{classify}(\text{final}(p)),\ \text{pred_CWE}($ where $\text{mitigations}(p) = \text{gen_mitigations}(\text{pred_CWE}(p))$ and $\text{Step11_link}(p)$ references the deep-dive attack-scenario analysis of OSCPM Ontology Step 11.

## 5.4 Cross-Layer Theorems

**Theorem 8** (Layer-Step Bijection). *There exists a structure-preserving bijection* $\beta : \{0, \ldots, 11\} \to \{0, \ldots, 11\}$ *between NBS-RASN layer indices and OSCPM ontology step indices such that, for every* $k$*, the axiom group* $\{\text{Ax}k.i\}_i$ *is the first-order formalisation of the ontological constraints of Step* $\beta(k)$*.*

*Proof.* $\beta$ is the identity on $\{0, \ldots, 11\}$. The mapping is verified layer by layer above: Layer 0 formalises Step 0 (input well-typedness); Layer 1 formalises Steps 1–2 (vocabulary anchoring and gatekeeper); Layer 2 formalises Steps 2–3 (causal chain); Layers 3–9 formalise Steps 3–9 in strict one-to-one correspondence; Layers 10–11 formalise Steps 10–11. The bijection is one-to-one and surjective by construction. □

**Theorem 9** (Score Decomposition Uniqueness). *For every valid* $p \in \mathcal{S}_\Pi$*, the decomposition* $\text{final}(p) = \text{det}(p) + \text{adj}(p)$ *is unique.*

*Proof.* $\text{det}(p)$ is uniquely determined by Ax5.7 given $O_4(p)$ (a sum of fixed-weight products with bounded factor scores). $\text{adj}(p)$ is uniquely determined by Ax7.7 given $(\ell_j(p))_j$ (a sum of single-valued functions on finite enumeration sorts). Their sum is unique; the clamp in Ax7.8 is deterministic. □

Theorem 12 (Appendix A) establishes input-output type safety: for all $k \in \{0, \ldots, 11\}$, if the input to Layer $k$ is well-typed in $\mathcal{L}(\Sigma)$, then the output is also well-typed, guaranteeing no type violations propagate through the network.

Table 3: Layer-Step Input-Output Type Summary ($\mathcal{P}(\cdot)$=power set; AmpDecomp=6-tuple of amplifier contributions)

| L | Name | Step | Input type $I_k$ | Output type $O_k$ |
|---|---|---|---|---|
| 0 | Input | 0 | external | $\mathcal{S}_\Lambda \times \mathbb{I}^4 \times \mathbb{B}^* \times \mathcal{P}(\mathcal{S}_C) \times$CatVec |
| 1 | Gatekeeper | 1–2 | $O_0$ | $\mathbb{B}$ |
| 2 | Causal Modeling | 2–3 | $O_0 \times O_1$ | $\mathcal{P}(\mathcal{S}_C)$ |
| 3 | Language Encoding | 3 | $\mathcal{S}_\Lambda$ | $\mathbb{I}$ |
| 4 | Factor Extraction | 4 | $O_0 \cup \{O_3\}$ | $\mathbb{I}^5$ |
| 5 | Det. Weighting | 5 | $\mathbb{I}^5$ | $\mathbb{I}$ |
| 6 | Project Filtering | 6 | $O_0 \cup O_3 \cup O_4$ | $\mathbb{B}$ |
| 7 | Expert Adjustment | 7 | $O_5 \times$ CatVec | $\mathbb{I}$ |
| 8 | Cross-Validation | 8 | $O_7 \times \mathcal{P}(\mathcal{S}_C)$ | $\mathbb{B}^5 \times \mathbb{R}$ |
| 9 | Representativeness | 9 | $\{O_6(p)\}_p$ | $\mathbb{B} \times \mathcal{P}(\mathcal{S}_O \cup \mathcal{S}_\Lambda)$ |
| 10 | Formatting | 10 | $O_7 \times O_2$ | $\mathbb{I} \times \mathcal{S}_R \times \mathcal{P}(\mathcal{S}_C) \times \mathcal{P}(\mathcal{S}_O) \times$AmpDecomp |
| 11 | Output | 11 | $O_{10}$ | $\mathbb{I} \times \mathcal{S}_R \times \mathcal{P}(\mathcal{S}_C) \times \mathcal{P}(\mathcal{S}_O) \times$Recs$\times$Scenarios |

### 5.5 Layer-Step Input-Output Summary

Table 4 summarises the formal input-output type at each layer/step, completing the cascading specification $\Sigma \rightarrow \mathcal{L}(\Sigma) \rightarrow \mathcal{T}(\Sigma, \mathcal{L})$.

## 6 Practical Validation: Assessing Cybersecurity Risks for 20 Open-Source Major Projects

### 6.1 Dataset and Project Selection

Twenty open-source projects were selected to span the entire design space: all 10 OWASP Top 10:2025 categories, all major language risk classes (C/C++, PHP, Java, JavaScript, Python, Go, Ruby), diverse deployment contexts, and at least one confirmed breach incident per project. Selection criteria included OWASP coverage (mandatory, at least one category), language class coverage (mandatory, represent a language class), incident anchoring (mandatory, at least one confirmed incident), prediction confidence floor (mandatory, $\geq 0.75$), and deployment diversity (preferred).

### 6.2 Factor Assessment Results

Table 1 shows the five factor scores for all 20 projects. Scores are on a 1-10 scale with anchored interpretations.

Table 4: Factor Scores for All Projects

| Project | lang | surface | eco | cve | sec |
|---|---|---|---|---|---|
| npm Ecosystem | 7 | 10 | 10 | 9 | 5 |
| WordPress + Plugins | 9 | 10 | 10 | 10 | 6 |
| Exim MTA | 10 | 9 | 2 | 10 | 5 |
| ImageMagick | 10 | 9 | 3 | 9 | 5 |
| Apache Log4j 2 | 7 | 8 | 8 | 9 | 6 |
| Jenkins | 7 | 9 | 9 | 9 | 6 |
| FFmpeg | 10 | 9 | 3 | 9 | 6 |
| OpenSSL | 10 | 9 | 7 | 8 | 7 |
| Redis | 10 | 8 | 4 | 7 | 7 |
| Apache Struts | 7 | 9 | 6 | 8 | 6 |
| Apache HTTP Server | 10 | 10 | 7 | 7 | 7 |
| Elasticsearch | 7 | 9 | 5 | 7 | 7 |
| PHPMailer | 9 | 8 | 5 | 7 | 6 |
| GitLab CE | 7 | 9 | 6 | 8 | 7 |
| Drupal | 9 | 9 | 7 | 7 | 7 |
| Kubernetes | 4 | 10 | 9 | 6 | 8 |
| Spring Framework | 7 | 9 | 8 | 7 | 7 |
| curl / libcurl | 10 | 7 | 6 | 7 | 8 |
| Keycloak | 7 | 8 | 5 | 6 | 7 |
| Flask + Jinja2 | 6 | 8 | 5 | 5 | 7 |

### 6.3 Expert Amplifier Assessment

Table 2 shows the amplifier level for each project across all six amplifiers. **Levels represent the marginal contribution level assigned during expert adjustment**, derived directly from the OSCPM ontology's per-project amplifier decompositions. An amplifier at its minimum level (NONE / SLOW / INCIDENTAL / INDIVIDUAL / LOW) contributes zero to the expert adjustment, indicating that the corresponding real-world property is already adequately captured by the five prediction factors. Non-minimum levels carry the contributions defined in the amplifier level table (Section 3.7).

Table 5: Amplifier Marginal-Contribution Levels per Project (derived from OSCPM ontology)

| Project | Blast | Prop | Struct | Default | Exploit | Critical |
|---|---|---|---|---|---|---|
| npm | CRITICAL | INSTANT | STRUCTURAL | NONE | INDIVIDUAL | LOW |
| WordPress | CRITICAL | SLOW | INCIDENTAL | NONE | INDIVIDUAL | LOW |
| Exim MTA | HIGH | SLOW | STRUCTURAL | FULL | MASS | LOW |
| ImageMagick | CRITICAL | SLOW | FOUNDATIONAL | PARTIAL | MASS | LOW |
| Log4j 2 | CRITICAL | INSTANT | FOUNDATIONAL | NONE | INDIVIDUAL | HIGH |
| Jenkins | NONE | SLOW | STRUCTURAL | PARTIAL | MASS | LOW |
| FFmpeg | HIGH | INSTANT | FOUNDATIONAL | NONE | INDIVIDUAL | LOW |
| OpenSSL | LOW | SLOW | INCIDENTAL | NONE | INDIVIDUAL | CRITICAL |
| Redis | LOW | SLOW | MIXED | FULL | MASS | LOW |
| Apache Struts | HIGH | SLOW | FOUNDATIONAL | NONE | MASS | LOW |
| Apache HTTPD | NONE | SLOW | INCIDENTAL | NONE | INDIVIDUAL | LOW |
| Elasticsearch | HIGH | SLOW | MIXED | FULL | MASS | LOW |
| PHPMailer | CRITICAL | FAST | STRUCTURAL | NONE | INDIVIDUAL | LOW |
| GitLab CE | HIGH | SLOW | STRUCTURAL | NONE | TARGETED | LOW |
| Drupal | LOW | SLOW | MIXED | NONE | INDIVIDUAL | LOW |
| Kubernetes | MEDIUM | SLOW | FOUNDATIONAL | NONE | MASS | HIGH |
| Spring Framework | MEDIUM | SLOW | MIXED | NONE | INDIVIDUAL | LOW |
| curl / libcurl | MEDIUM | SLOW | INCIDENTAL | NONE | INDIVIDUAL | HIGH |
| Keycloak | HIGH | SLOW | STRUCTURAL | NONE | INDIVIDUAL | CRITICAL |
| Flask + Jinja2 | HIGH | INSTANT | FOUNDATIONAL | NONE | INDIVIDUAL | LOW |

### 6.4 Final Scored Results

Table 3 shows the deterministic scores, expert adjustments, and final scores.

Table 6: Final Scores and Risk Levels

| Project | det_score | Expert Adj | final_score | Risk Level |
|---|---|---|---|---|
| npm Ecosystem | 8.55 | +0.95 | 9.50 | CRITICAL |
| WordPress + Plugins | 9.35 | +0.05 | 9.40 | CRITICAL |
| Exim MTA | 7.65 | +1.65 | 9.30 | CRITICAL |
| ImageMagick | 7.65 | +1.55 | 9.20 | CRITICAL |
| Apache Log4j 2 | 7.75 | +1.35 | 9.10 | CRITICAL |
| Jenkins | 8.20 | +0.80 | 9.00 | CRITICAL |
| FFmpeg | 7.75 | +1.25 | 9.00 | CRITICAL |
| OpenSSL | 8.45 | +0.35 | 8.80 | HIGH |
| Redis | 7.40 | +1.30 | 8.70 | HIGH |
| Apache Struts | 7.40 | +1.20 | 8.60 | HIGH |
| Apache HTTP Server | 8.50 | +0.00 | 8.50 | HIGH |
| Elasticsearch | 7.10 | +1.40 | 8.50 | HIGH |
| PHPMailer | 7.25 | +1.15 | 8.40 | HIGH |
| GitLab CE | 7.50 | +0.80 | 8.30 | HIGH |
| Drupal | 8.00 | +0.30 | 8.30 | HIGH |
| Kubernetes | 7.30 | +0.90 | 8.20 | HIGH |
| Spring Framework | 7.70 | +0.30 | 8.00 | HIGH |
| curl / libcurl | 7.65 | +0.35 | 8.00 | HIGH |
| Keycloak | 6.65 | +1.15 | 7.80 | MEDIUM |
| Flask + Jinja2 | 6.20 | +1.30 | 7.50 | MEDIUM |

### 6.5 Detailed Score Decomposition

The key innovation of the NBS-RASN is that every score is fully decomposable. For example, consider the highest-risk project, npm Ecosystem:

**npm Ecosystem Score Decomposition:**

$$det_score = lang(7) \times 0.25 + surface(10) \times 0.25 + eco(10) \times 0.20 + cve(9) \times 0.20 + sec(5) \times 0.10$$

$$det_score = 1.75 + 2.50 + 2.00 + 1.80 + 0.50 = 8.55$$

$$expert_adjustment = +0.95 = blast_radius(CRITICAL) \to +0.50 + propagation(INSTANT) \to$$

$$\to +0.25 + structural(STRUCTURAL) \to +0.20$$

$$final_score = 9.50$$

This decomposition enables security teams to understand exactly why npm received a CRITICAL score: the deterministic factors capture the baseline risk, while the expert adjustment adds additional weight for the critical blast radius, instant propagation, and structural nature of the supply chain risk.

### 6.6 Cross-Validation Results

Cross-validation against CISA KEV, NVD, and breach history yielded: precision 0.82–0.95 (target $\geq$ 0.80, confirmed), recall 0.71–0.88 (target $\geq$ 0.70, confirmed), Spearman $\rho$ 0.72–0.84 (target $\geq$ 0.70, confirmed), temporal holdout 0.11–0.17 (target $\leq$ 0.20, confirmed), and expert agreement ICC 0.78–0.89 (target $\geq$ 0.75, confirmed). All metrics met their targets, validating the architecture's predictive capability.

**Note on confidence values.** Each per-project confidence score (0.79–0.97) is a precision-recall-derived certainty estimate computed through bootstrap validation against confirmed CVE incidents in the CISA Known Exploited Vulnerabilities (KEV) catalog and NVD. These values represent empirical confirmation frequencies — the proportion of structural predictions confirmed by real-world breach and incident data — and are not softmax-style internal output probabilities from a trained classifier.

## 7 Key Original Contributions

Our theoretical contributions include the formalization of a hybrid Neuro-Bayesian-Symbolic Residual Attention Shallow Network combining neural processing, Bayesian reasoning, symbolic reasoning, residual learning, and attention mechanisms. We demonstrate that encoding five epistemological axioms as a gatekeeper layer produces predictions that are semantically valid, causally grounded, scientifically falsifiable, completely decomposable, and fully transparent. We show that reasoning depth can substitute for layer depth, challenging the assumption that deep learning requires deep networks. We introduce data-control separation architecture where data connections transport values and control signals gate propagation.

Our practical contributions include a complete cybersecurity risk assessment system that scores any open-source project on a 0-10 scale with risk-level classification, provides fully decomposable scores, maps to CWE and OWASP standards, specifies falsification conditions, and declares all hidden parameters. We validate on 20 major open-source projects spanning all OWASP categories and language risk classes, with per-project confidence scores (0.79-0.97) calibrated through bootstrap validation and epistemic uncertainty classification.

Our methodological contributions include residual attention with six interpretable heads with pre-defined semantics, bounded contributions, and full decomposition, and feedback learning without backpropagation through empirical validation, precision/recall targets, rank correlation goals, temporal holdout validation, and expert agreement.

## 8 Discussion

The finding that a shallow network (12 layers, 80 neurons) can achieve deep learning characteristics has significant implications for explainable AI. Conventional wisdom holds that model performance scales with layer depth. Our results suggest that reasoning depth can substitute for layer depth when domain knowledge is explicit. The NBS-RASN performs 11 reasoning steps, each corresponding to a stage of human expert reasoning. This suggests that in well-understood domains, explicit reasoning can be more efficient than implicit learning through many layers.

The architecture demonstrates that explainability need not be an afterthought (post-hoc explanation) but can be built into the network structure itself (ante-hoc explanation). By designing every neuron with a named semantic function and every layer with a specific reasoning step, the network's reasoning is transparent by construction. This contrasts with post-hoc methods like LIME and SHAP, which may not faithfully represent model reasoning.

Limitations include the limited dataset of 20 curated projects, subjective expert adjustment requiring expert judgment (though amplifiers are named and bounded), and the feedback loop currently gated by human review. Future work should include scaling to hundreds of projects, training neural networks to predict amplifier levels, automating feedback learning, incorporating temporal dynamics, and addressing coverage gaps in A09, Ruby-primary projects, and Rust-language projects.

# 9 Conclusion

We have presented the Neuro-Bayesian-Symbolic Residual Attention Shallow Network, a novel hybrid architecture for explainable cybersecurity risk evaluation. The architecture demonstrates that deep learning characteristics can be achieved within a shallow network through explicit reasoning depth substituting for layer depth, residual attention mechanisms enabling non-linear corrections, feedback-based learning enabling continuous improvement, and epistemological axioms as hard constraints ensuring all predictions are scientifically valid.

The architecture produces fully decomposable risk scores, with every component traceable to named, interpretable neurons. Validation on 20 open-source projects spanning all OWASP categories and language risk classes demonstrates prediction confidence between 0.79 and 0.97, with all cross-validation metrics meeting or exceeding targets.

This work addresses the widely held association between network depth and representational power [22, 23], showing that a shallow network with deep reasoning can outperform opaque models in high-stakes cybersecurity domains where interpretability is non-negotiable. We believe this architecture establishes a new architectural pattern within the explainable AI paradigm, demonstrating that accuracy and explainability need not be in conflict in critical application domains.

## Funding

Current research was partially supported by Spiru Haret of Bucharest, Romania

## Author contributions

Nicolaie Popescu-Bodorin led the conceptualization and formal design of the NBS-RASN architecture, including the neuro-Bayesian-symbolic framework, the 12-layer/80-neuron specification, the gatekeeper axiom mechanism, the residual attention and feedback-learning formulation, and the theoretical foundations and proofs. He was responsible for methodology, formal specification, software/implementation design, visualization, project administration, and preparation of this manuscript.

Madeleine Togher contributed cybersecurity domain expertise and validation, including the OWASP Top 10:2025 and CWE mapping, project selection and representativeness assessment, curation and interpretation of CISA KEV/NVD evidence, cross-validation and expert-judgment calibration, and critical review.

Both authors read and approved the final manuscript.

## Conflict of interest

The authors declare no conflict of interest.

## Supplementary Materials

All companion artefacts for this article are openly available in the corresponding NBS-RASN-Explainable-Deep-Learning-for-Cybersecurity-Risk-Assessment github repository [25] that contains the following six companion files, each serving a distinct and complementary role in the complete documentation of the NBS-RASN architecture.

**1. OSCPM v1.0 Ontology** (`017.oscpm_ontology_v1.xml / .owl`). The formal ontological knowledge base that underlies the entire NBS-RASN architecture, encoded as an OWL/XML document. The ontology declares the 13 sorts ($\mathcal{S}_\Lambda$, $\mathcal{S}_C$, $\mathcal{S}_O$, $\mathcal{S}_B$, $\mathcal{S}_P$, $\mathcal{S}_S$, $\mathcal{S}_D$, $\mathcal{S}_E$, $\mathcal{S}_I$, $\mathcal{S}_R$, $\mathcal{S}_\Pi$, $\mathcal{I}$, $\mathcal{W}$), all 7 language-risk constants ($\rho$), all 5 factor weights, all 6 amplifier functions with their normative level maxima (the ceilings defined in Table 2), all 75 formal axioms (Ax0.1–Ax11.6) organized in 12 axiom groups (Ax0–Ax11, one per network layer), and all predicates and functions defined in Section 5. The OSCPM ontology is the definitive reference for the OSCPM v1.0 framework referenced throughout this article.

**2. Formal Architecture Specification** (`017.NBS_RASN_Formal_Architecture_Spec.txt`). A machine-readable and human-readable plain-text specification of the complete NBS-RASN architecture in a structured keyword-property format. The file documents: all 80 neurons (with layer, label, output sort, and governing axiom for each), all 195 directed edges organized into 32 named edge groups and classified by type (172 DATA, 11 CTRL, 1 RESIDUAL, 6 FEEDBACK, 5 SKIP), all 12 layer transforms with their cascade flow, all special flows (RESIDUAL, FEEDBACK, CONTROL, SKIP), all 75 axioms, the full vocabulary ($\Sigma$), the formal language $\mathcal{L}(\Sigma)$, and seven machine-checkable consistency invariants including the edge-count summary and the amplifier ceiling constraint ($\forall a, p :\ \text{contrib}(a, p) \leq \text{level_max}(a)$). This file is the authoritative source for programmatic consumers of the architecture.

**3. Interactive Network Graph** (`017.NBS_RASN_network_graph.html`). A self-contained, single-file HTML/SVG interactive visualization of the NBS-RASN as a layered directed graph. All 80 neurons are rendered as labeled circles positioned in 12 vertical columns (L0–L11); all 195 connections are drawn with distinct visual encodings for each edge type (DATA: thin grey, CTRL: amber dashed, RESIDUAL: thick red, FEEDBACK: purple arc, SKIP: green dashed). Clicking any neuron highlights all its incident connections and displays the neuron's governing axiom in full. The file requires no server, no installation, and no internet connection; it opens directly in any modern browser. It is designed to serve as a companion figure for presentations and reviews.

**4. Propagation Simulator** (`017.NBS_RASN_Propagation_Simulator.html`). An animated, step-by-step execution-trace simulator for all 20 validated projects. For each project the simulator animates the layer-by-layer forward pass through all 13 simulation steps (project selection through L11 output), displaying computed intermediate values on neurons, edge activation flows, amplifier contributions, the deterministic score det($p$), the expert adjustment adj($p$), and the final score final($p$) with risk-level classification. The simulator supports both fully automatic mode (adjustable speed) and manual step-by-step mode. All 20 per-project datasets embedded in the file are consistent with Table 3 of this article. The file is self-contained and opens directly in any modern browser with no dependencies.

**5. Coq Proof Certificates** (`017.NBS_RASN_Proofs.v`). Machine-checked formal proof certificates for Theorems 10, 11, and 12 (Appendix A), compiled under Coq 8.17 / Rocq Prover. All three theorems are proved without any `Admitted` leaf: Theorem A.1 (Axiom-Induced Falsifiability) is proved by constructing explicit witness networks that violate the Gatekeeper Constraint (GK) and Data-Control Separation (DCS) properties; Theorem A.2 (Score Decomposition Uniqueness) is proved by fold-left extensionality over `all_layers`; Theorem A.3 (Input-Output Type Safety) is proved by induction on the layer list using the `layer_preserves_wts` architecture invariant. The file also certifies four structural invariants about `all_layers`: length 12, no duplicates (`NoDup`), first element L0, last element L11. Compile and verify with: `coqc 017.NBS_RASN_Proofs.v`.

**6. Python Architecture Specification** (`NBS_RASN_arch.py`). A zero-external-dependency Python module that encodes the full NBS-RASN architecture as a computational specification artefact. The module contains: (i) the complete neuron catalogue (80 entries, each with layer, label, output sort, and governing axiom); (ii) the complete edge list (195 entries, each with source, target, and edge type); (iii) axiom enforcement functions `check_gk()` and `check_dcs()` implementing the structural validity predicate sv_bool from Theorem A.1; (iv) amplifier table definitions with normative level maxima as defaults, consistent with the `level_max` values in the OSCPM ontology (Companion File 1); (v) a `forward_pass()` function that computes det($p$), adj($p$), and final($p$) with per-layer score decomposition and type-safety assertions; and (vi) a module-level self-test that prints `Architecture self-certification: PASSED` on successful import or direct execution (`python NBS_RASN_arch.py`). The module depends only on Python's standard library.

# A Supporting Theorems

The following theorems are referenced in Sections 4.1 and 5. Their proofs are collected here to maintain narrative flow in the main text.

**Theorem 10** (Axiom-Induced Falsifiability). *Every prediction produced by an NBS-RASN is falsifiable.*

*Proof.* By Axiom A3 (Section 3.1), every prediction must specify its falsification condition: which CWE classes are predicted, and what evidence would confirm or refute the prediction. The gatekeeper (Definition 2) enforces that this condition exists before propagation continues. Therefore, no NBS-RASN prediction is scientifically unfalsifiable. □

**Theorem 11** (Complete Transparency). *Every parameter influencing NBS-RASN scoring is declared in a manifest.*

*Proof.* By Axiom A5 (Section 3.1), the gatekeeper verifies that all parameters are declared in the hidden-parameters-manifest before propagation continues. The manifest enumerates all 11 hidden parameters (prediction-confidence, epistemic-uncertainty, lang-surface-synergy, active-exploitation-premium, structural-vs-incidental-ratio, patch-lag, representativeness-criterion, causal-query-framing, falsifiability-threshold, temporal-validity-window, and recalibration-trigger), each tagged with its step-origin, type, range, and calibration method. Therefore, the NBS-RASN exhibits complete parameter transparency. □

**Theorem 12** (Input-Output Type Safety). *For all $k \in \{0, \ldots, 11\}$, if the input to Layer $k$ is well-typed in $\mathcal{L}(\Sigma)$, then the output is well-typed in $\mathcal{L}(\Sigma)$.*

*Proof.* Each layer function $F_k : I_k \to O_k$ is specified by axioms that are well-typed under $\mathcal{L}(\Sigma)$. Type safety follows by induction on $k$: the base case ($k = 0$) is guaranteed by Ax0.1–Ax0.3; the inductive step holds because each $F_k$ maps well-typed inputs to well-typed outputs as witnessed by the axiom group at Layer $k$. □

# References

[1] Gunning, D.; Stefik, M.; Choi, J.; Miller, T.; Stumpf, S.; Yang, G.-Z. (2019). XAI – Explainable artificial intelligence, *Science Robotics*, 4(37), eaay7120.

[2] Guidotti, R.; Monreale, A.; Ruggieri, S.; Turini, F.; Giannotti, F.; Pedreschi, D. (2018). A survey of methods for explaining black box models, *ACM Computing Surveys*, 51(5), 1-42.

[3] Sommestad, T.; Holm, H.; Ekstedt, M. (2012). Estimates of success rates of remote arbitrary code execution attacks, *Information Management & Computer Security*, 20(2), 107–122. https://doi.org/10.1108/09685221211235625

[4] FIRST (2019). Common Vulnerability Scoring System (CVSS) v3.1. Available: https://www.first.org/cvss/

[5] MITRE (2025). CWE Top 25 Most Dangerous Software Weaknesses 2025. Available: https://cwe.mitre.org/top25/

[6] OWASP (2025). OWASP Top 10:2025. Available: https://owasp.org/Top10/2025/

[7] Morrison, P.; Moye, D.; Williams, L. (2014). Mapping the field of software security metrics, North Carolina State University, Dept. of Computer Science, Technical Report TR-2014-14.

[8] Ghaffarian, S.M.; Shahriari, H.R. (2017). Software vulnerability analysis and discovery using machine-learning and data-mining techniques: A survey, *ACM Computing Surveys*, 50(4), 1-36.

[9] Li, Z.; Zou, D.; Xu, S.; Ou, X.; Jin, H.; Wang, S.; Deng, Z.; Zhong, Y. (2018). VulDeePecker: A deep learning-based system for vulnerability detection, *arXiv preprint arXiv:1801.01681.*

[10] Zeng, P.; Lin, G.; Pan, L.; Tai, Y.; Zhang, J. (2020). Software vulnerability analysis and discovery using deep learning techniques: A survey, *IEEE Access*, 8, 197158-197172.

[11] Ribeiro, M.T.; Singh, S.; Guestrin, C. (2016). "Why should I trust you?" Explaining the predictions of any classifier, In *Proc. 22nd ACM SIGKDD*, 1135-1144.

[12] Lundberg, S.M.; Lee, S.-I. (2017). A unified approach to interpreting model predictions, In *Advances in Neural Information Processing Systems*, 30, 4765-4774.

[13] Shrikumar, A.; Greenside, P.; Kundaje, A. (2017). Learning important features through propagating activation differences, In *Proc. 34th ICML*, 3145-3153.

[14] Ras, G.; van Gerven, M.; Haselager, P. (2018). Explanation methods in deep learning: Users, values, concerns and challenges, In *Explainable and Interpretable Models in Computer Vision and Machine Learning*, 19-36.

[15] Adebayo, J.; Gilmer, J.; Muelly, M.; Goodfellow, I.; Hardt, M.; Kim, B. (2018). Sanity checks for saliency maps, In *Advances in Neural Information Processing Systems*, 31, 9505-9515.

[16] d'Avila Garcez, A.; Besold, T.R.; Lamb, L.C.; Hitzler, P. (2015). Neural-symbolic learning and reasoning: Contributions and challenges, In *Proc. AAAI Spring Symposium.*

[17] Balog, M.; Gaunt, A.L.; Brockschmidt, M.; Nowozin, S.; Tarlow, D. (2017). DeepCoder: Learning to write programs, In *Proc. ICLR.*

[18] Murdoch, J.W.; Singh, C.; Kumbier, K.; Abbasi-Asl, R.; Yu, B. (2019). Definitions, methods, and applications in interpretable machine learning, *Proceedings of the National Academy of Sciences*, 116(44), 22071-22080.

[19] Schölkopf, B. et al. (2021). Toward causal representation learning, *Proceedings of the IEEE*, 109(5), 612-634.

[20] He, K.; Zhang, X.; Ren, S.; Sun, J. (2016). Deep residual learning for image recognition, In *Proc. CVPR*, 770-778.

[21] Vaswani, A. et al. (2017). Attention is all you need, In *Advances in Neural Information Processing Systems*, 30, 5998-6008.

[22] LeCun, Y.; Bengio, Y.; Hinton, G. (2015). Deep learning, *Nature*, 521(7553), 436-444.

[23] Goodfellow, I.; Bengio, Y.; Courville, A. (2016). *Deep Learning*, MIT Press.

[24] Zheng, Z.; Zhang, B.; Liu, Y.; Ren, J.; Zhao, X.; Wang, Q. (2022). An approach for predicting multiple-type overflow vulnerabilities based on combination features and a time series neural network algorithm, *Computers & Security*, 114, 102572.

[25] Popescu-Bodorin, N.; Togher, M. (2026). NBS-RASN: Explainable Deep Learning for Cybersecurity Risk Assessment — Companion Repository. Available: https://github.com/bodorin/NBS-RASN-Explainable-Deep-Learning-for-Cybersecurity-Risk-Assessment/